\documentclass[runningheads]{llncs}
\usepackage[T1]{fontenc}
\usepackage{graphicx}
\usepackage{xcolor}
\usepackage{subcaption}
\usepackage{algorithm2e}
\usepackage{amsmath}
\usepackage{amssymb}
\newtheorem{result}{Result}

\usepackage{comment}

\begin{document}
\title{A Learning-Based Hybrid Decision Framework for Matching Systems with User Departure Detection}
\titlerunning{Learning-Based Hybrid Matching Framework}
% If the paper title is too long for the running head, you can set
% an abbreviated paper title here
%

\author{Ruiqi Zhou\inst{1}\orcidID{0009-0006-8220-6265} \and
Donghao Zhu\inst{2,3}\orcidID{0000-0002-1889-3076} \and
Houcai Shen\inst{1}\orcidID{0000-0003-1704-9129}}

\authorrunning{R. Zhou et al.}
% First names are abbreviated in the running head.
% If there are more than two authors, 'et al.' is used.
%
\institute{School of Management and Engineering, Nanjing University,  China
\email{zrq6@smail.nju.edu.cn, hcshen@nju.edu.cn}
\and
The Institute of Statistical Mathematics, Tokyo, Japan \email{donghao.zhu9407@gmail.com} \and
The University of Tokyo Market Design Center, Tokyo, Japan.
}
\maketitle              % typeset the header of the contribution
\begin{abstract}
In matching markets such as kidney exchanges and freight exchanges, delayed matching has been shown to improve overall market efficiency. 
The benefits of delay are highly sensitive to participants’ sojourn times and departure behavior, and delaying matches can impose significant costs, including longer waiting times and increased market congestion.
These competing effects make fixed matching policies inherently inflexible in dynamic environments.
We propose a learning-based $\mathsf{Hybrid}$ framework that adaptively combines immediate and delayed matching. 
The framework continuously collects data on user departures over time, estimates the underlying departure distribution via regression, and determines whether to delay matching in the subsequent period based on a decision threshold that governs the system’s tolerance for matching efficiency loss.
The proposed framework can substantially reduce waiting times and congestion while sacrificing only a limited amount of matching efficiency. 
By dynamically adjusting its matching strategy, the $\mathsf{Hybrid}$ framework enables system performance to flexibly interpolate between purely greedy and purely patient policies, offering a robust and adaptive alternative to static matching mechanisms.
\end{abstract}

\section{Introduction}

Online matching problems arise in a wide range of economic and operational settings, including labor markets, kidney exchanges, freight platforms, and online service systems.
A central challenge in these environments is that matching decisions must be made over time under uncertainty, as participants arrive and depart dynamically.
A growing body of literature has documented that the performance of online matching policies, particularly the benefits of delaying matches to thicken the market, is highly sensitive to the distribution of participants’ sojourn times and departure behavior\cite{akbarpour2020thickness,baumler2022superiority,kakimura2021dynamic}.
Even when the mean departure time is identical, differences in distributional shape can reverse the relative performance of matching policies.
When a substantial fraction of agents has long sojourn times, immediate matching tends to perform better.
Conversely, when most agents depart quickly and the mean is inflated by a small fraction of long-stayers, delaying matches to thicken the market yields superior outcomes.
\cite{baumler2022superiority} attributes this difference to the fact that immediate matching ignores the opportunity for agents to be matched while waiting, whereas maintaining a large pool through delayed matching can substantially improve matching efficiency.

A canonical example motivating our setting is kidney exchange~\cite{akbarpour2020thickness,ashlagi2021kidney,baumler2022superiority,carvalho2021robust,dickerson2017multi}, where each agent represents a patient–donor pair, and a feasible match corresponds to a mutually compatible exchange between two pairs.
Due to the medical rarity of physiological compatibility, a newly arrived agent is compatible with any given agent in the pool only with a very small probability.
As a result, the matching environment is inherently sparse but characterized by high match value~\cite{kakimura2021dynamic}.
In this setting, a newly arrived agent is not guaranteed to have any feasible matching candidate upon arrival, and when feasible candidates exist, their number is typically small.
Agents do not rank or strategically evaluate candidates; instead, compatibility is exogenously determined and matches are formed whenever a feasible pair is identified.
Because agents incur a high cost of waiting and leave the market if unmatched after a limited time, the opportunity to match is scarce and valuable, making patience a critical design consideration in the matching policy.

Beyond matching efficiency, delayed matching introduces additional operational costs, most notably increased waiting times for participants and higher system congestion.
These costs are particularly salient in application domains such as healthcare systems, where prolonged waiting may directly affect patient outcomes and resource utilization.
In such settings, matching decisions are often mediated by human–machine interactions: 
decision support systems rely on continuously updated patient information, such as medical records, test results, and risk assessments, to guide operational choices in real time.
This raises a fundamental question: 
\textit{How should a matching system leverage evolving user information to adapt its matching strategy to changing conditions, rather than committing to a fixed policy?}
We propose a learning-based decision framework, termed $\mathsf{Hybrid}$, for dynamic matching systems that explicitly models users’ sojourn times and departure behavior.
Recognizing that immediate matching can substantially reduce waiting time and congestion despite incurring limited efficiency loss, the framework continuously collects departure data, estimates the underlying distribution using statistical learning techniques, and adaptively decides whether to delay matching based on a matching efficiency loss tolerance threshold.
This design enables dynamic and time-varying policy selection that balances efficiency and operational costs.

The $\mathsf{Hybrid}$ framework is flexible by design, dynamically combining immediate and delayed matching rather than committing to either extreme.
We show that the proposed framework can significantly reduce average waiting times and system congestion while sacrificing only a small amount of matching efficiency.
By allowing system performance to interpolate smoothly between purely greedy and purely patient policies, our approach offers an alternative to static matching mechanisms.
More broadly, the framework provides a general decision-making paradigm for learning-driven control in online matching systems, opening avenues for future research on adaptive, data-informed matching policies.

The remainder of the paper is organized as follows.
Section 2 reviews the related works.
Section 3 introduces the proposed decision framework.
Section 4 presents the matching models and policy methodologies to which the framework can be applied.
Section 5 reports numerical experiments and examines the impact of key control variables.
Finally, Section 6 concludes the paper and discusses several open questions for future research.

\section{Related Work}

Our research relates to three areas of literature: healthcare operational frameworks, decision-theoretic methodologies, and matching policies and algorithms. Transplantable organs are often scarce, making organ matching an important research area within the framework and mechanisms of healthcare operations. Organ allocation refers to assigning organs to users on a list \cite{bertsimas2013fairness}; exchange refers to two mismatched patient-donor pairs exchanging kidneys \cite{ashlagi2021kidney,papalexopoulos2024reshaping}. Circular exchange involves several mismatched patient/donor pairs, and to ensure fairness, several surgeries are typically performed simultaneously \cite{ashlagi2021kidney,carvalho2021robust}. Chained exchange: This type of kidney exchange is usually initiated by a nondirected donor, who donates to the first patient in the chain, whose donor then donates to the second patient pair, and so on \cite{ashlagi2021kidney}. Dickerson and Sandholm \cite{dickerson2017multi} proposed a cross-organ donation scheme, where kidneys and livers could be bartered for each other. Our research is also related to the trade-off between fairness and efficiency in healthcare \cite{ashlagi2021kidney,bertsimas2013fairness,papalexopoulos2024reshaping}, and issues of ethics \cite{guidolin2022ethical}. To address potential partial failures during the exchange process, a solution is provided that minimizes the material and psychological costs associated with changing the allocation \cite{carvalho2021robust}. In addition, other matching and operational issues in healthcare include bed assignment \cite{demeester2010hybrid}, nurse allocation \cite{morse2024centralized}, and so on. Most existing operational frameworks address static problems. Building upon these theoretical and practical foundations, we have constructed a dynamic decision-making framework. This framework combines learning and optimization methods to balance multiple objectives.

Recent research aims to improve the predict-then-optimize paradigm by integrating operational decision into the estimation of unknown parameters. Known variously as smart predict then optimize \cite{elmachtoub2022smart,kao2009directed}, decision-focused learning \cite{mandi2024decision}, end-to-end optimization \cite{donti2017task}, or integrated learning and optimization \cite{sadana2025survey}. The operational data analytics (ODA) framework \cite{feng2023framework} examines a specific subset of these data-driven problems characterized by a homogeneity condition that links the scaling of objectives, parameters, and decisions. Our research is also grounded in general statistical theories of parameter estimation and selection~\cite{warwick2005choosing}. 

Regarding online (dynamic) matching markets, the most commonly considered type is a bipartite matching market, where the nodes are divided into two sets, and nodes in one set can only be matched with nodes in the other set. Research on this topic includes both one-sided online matching and fully online matching. Our framework, however, considers dynamic matching in a non-bipartite setting. Recent works debate the sufficiency of greedy policies versus batched interventions. Kerimov et al. \cite{kerimov2025optimality} establish that greedy policies are hindsight optimal in two-way networks. For general multi-way markets, Kerimov et al. \cite{kerimov2024dynamic} propose periodic clearing to balance short- and long-term value, arguing that delays are necessary for constant regret.  Gupta \cite{gupta2024greedy} demonstrated that simple greedy algorithms can achieve bounded regret ($O(\frac{1}{\epsilon})$) in multiway settings without periodic resolving. 

\section{Framework}
\begin{figure}[tbp]
\centering
\includegraphics[width=1\textwidth]{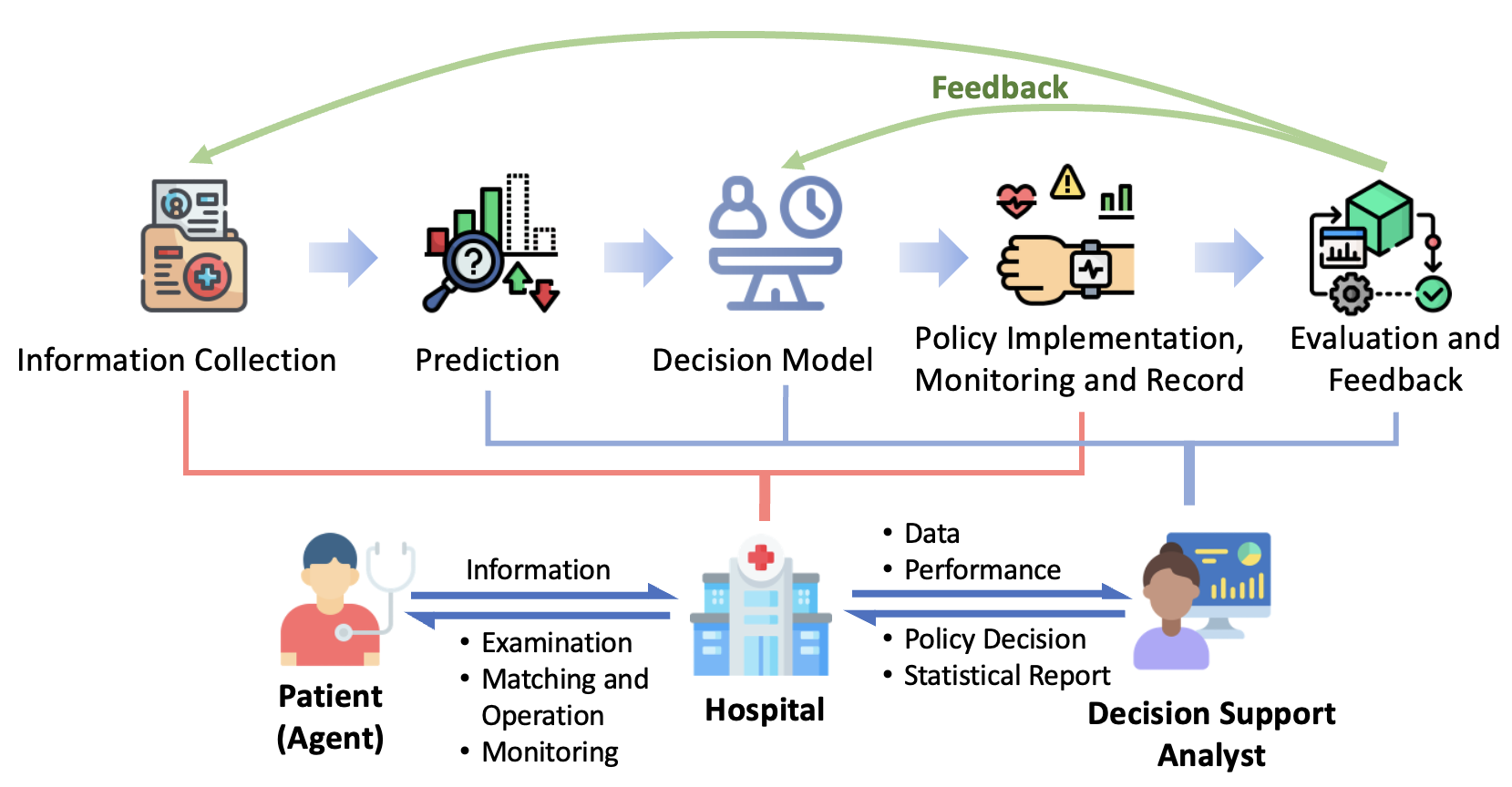}
\caption{Decision framework: a kidney exchange example.}\label{framework}
\end{figure}

The framework is designed as a data-driven adaptive matching system that continuously evolves its matching policy through interaction and feedback over time.
As illustrated in Figure~\ref{framework}, the system workflow consists of five interconnected stages: 
\textit{Information Collection}, \textit{Prediction}, \textit{Decision Model}, \textit{Policy Implementation}, \textit{Monitoring and Record}, and \textit{Evaluation and Feedback}. 
Rather than operating as a one-shot pipeline, these stages are linked through feedback loops that enable ongoing adjustment and learning, allowing the system to adapt its matching behavior to changing environments.
Specifically, within each time window, the system first gathers departure information from agents and the operational platform during the information collection stage. 
This individual-level departure information is then processed in the prediction stage to estimate the aggregate departure time distribution over the corresponding time window.
The predicted outcomes are subsequently provided to the decision model, which selects a preferred matching policy for the next time window.
The selected policy is executed in the policy implementation, monitoring and record stage, where operational outcomes and performance indicators are continuously monitored and logged.
Finally, in the feedback stage, the system assesses the results of the matching process and feeds performance insights back to earlier stages to support adaptive policy adjustment.

\begin{figure}[t]
\centering
\includegraphics[width=1\textwidth]{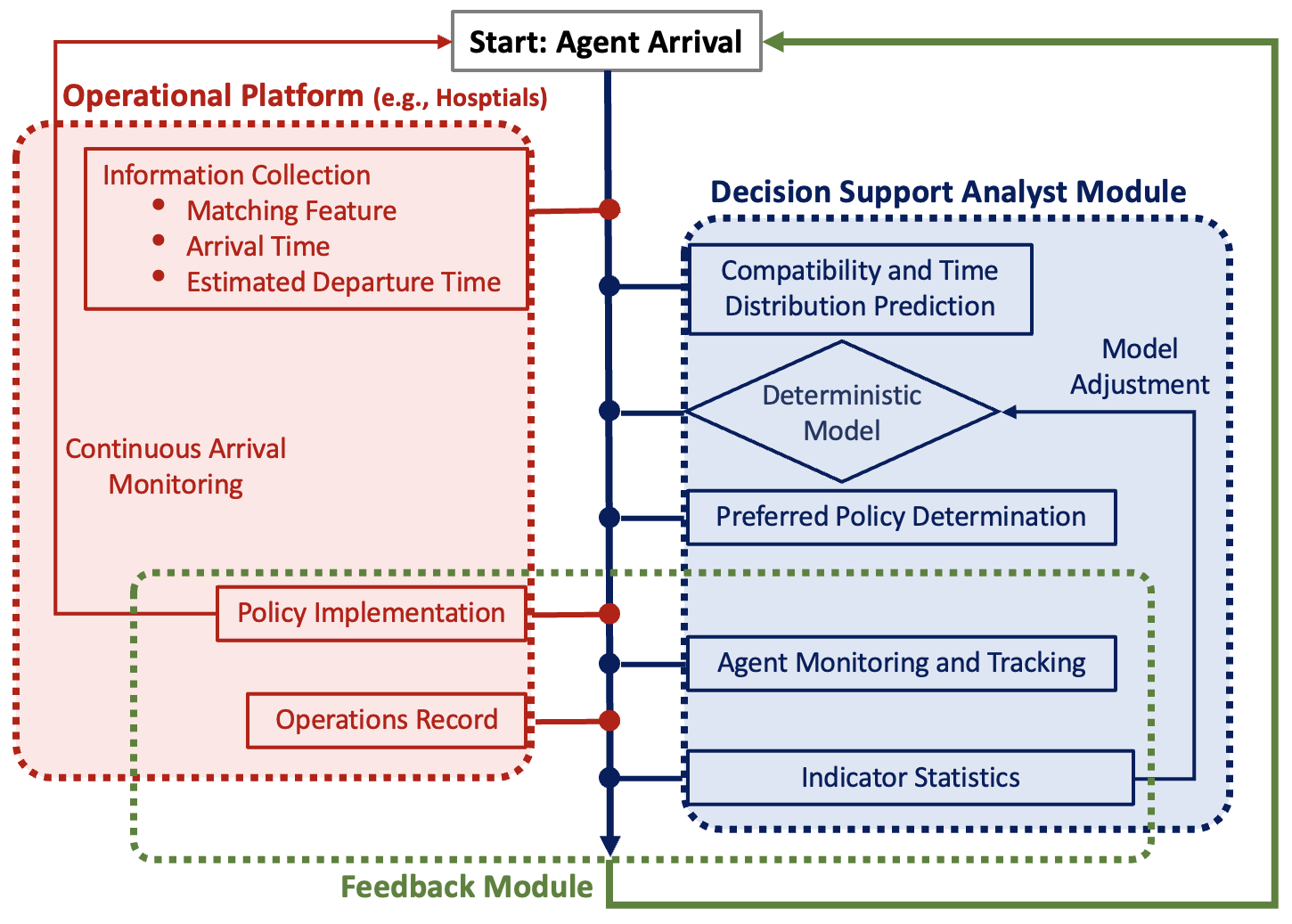}
\caption{A data-driven adaptive matching system.}\label{procedure}
\end{figure}

The framework distinguishes the roles and interactions among three participating modules. 
The \textit{Patient} (serves as the user, or agent) acts as both an information provider and a service recipient, engaging in continuous interaction with the system. 
The \textit{Hospital Module}, serving as the operational platform, is responsible for executing matching decisions, monitoring system operations, and interacting directly with users. 
The \textit{Decision Support Analyst Module} functions as the analytical and intelligent core of the system, leveraging collected data and performance feedback to generate policy decisions and statistical reports. 
The system aggregates data over a defined temporal window before initiating its predictive and decision-making functions.
Through structured information flows and feedback mechanisms, these entities together form a tightly coupled human–system collaborative loop.
From an interaction-centered perspective, the framework emphasizes not only the selection of matching policies, but also how the system learns, adapts, and refines its decision behavior through repeated human–system interaction, enabling robust and interpretable decision support in dynamic matching environments. 
Figure~\ref{procedure} provides a flowchart-based illustration of the detailed workflows and interactions among the individual modules.

\subsection{The Operational Platform}
The hospital module functions as the operational front end of the decision framework, serving as the primary interface between the system and incoming agents. 
As illustrated in the red box in Figure~\ref{procedure}, upon each agent’s arrival, the module collects essential information, such as matching-relevant attributes, the arrival timestamp, and an estimated departure time, which forms the foundational input for downstream prediction and decision modeling. 
The module continuously monitors new arrivals to maintain an accurate, real-time representation of the active matching pool. 
Based on the policy selected by the decision support analyst module, the hospital module executes the prescribed matching action, whether this entails performing an immediate match or deferring a decision for future opportunities. 
All operational events, including arrivals, executed matches, and agent departures, are systematically recorded, providing the empirical data required for subsequent system evaluation and feedback.

\subsection{The Decision Support Analyst Module}
The decision support analyst module, illustrated in the blue box in Figure~\ref{procedure}, serves as the analytical core of the framework and is responsible for transforming raw operational data into actionable matching decisions. 
Upon receiving updated information from the hospital module, it first predicts compatibility profiles and models agents’ estimated departure-time distributions, using statistical tools (e.g., log-normal models) to capture heterogeneous leaving behaviors. 
These predictions feed into a deterministic decision model that computes the parameter thresholds governing the optimal matching strategy, thereby identifying whether a patient or greedy policy is preferable under current market conditions. 
In parallel, the module continuously monitors the evolving status of all active agents and compiles key performance indicators, including match rates, waiting times, and departure-related losses, to assess system behavior in real time. 
These performance statistics inform the module’s model-adjustment mechanism, allowing it to refine its predictive and decision models over time and ensuring that the matching policy remains adaptive to changing market dynamics.

\subsection{The Feedback Module}
The feedback module, shown in the green box of Figure~\ref{procedure}, provides the mechanism that links the operational actions of the hospital module with the analytical updates of the decision support analyst module, forming a closed-loop control system. 
After each matching decision is executed, the resulting operational records, such as realized matches, waiting times, departure outcomes, and other performance indicators, are collected and fed into the evaluation process. 
Using these data, the system computes statistical measures that quantify the effectiveness of the implemented policy. 
These performance indicators are then returned to the analytical module, where they trigger model adjustment procedures, allowing the prediction models and decision thresholds to be recalibrated whenever systematic deviations or emerging patterns are detected. 
Through this continuous feedback cycle, the framework dynamically adapts its decision variables across time windows, ensuring that both prediction accuracy and policy optimality remain aligned with the evolving market environment.

During the initial stage of system operation, the framework defaults to the $\mathsf{Patient}$ policy as a safeguard against data scarcity.
While the $\mathsf{Patient}$ policy may incur higher waiting costs, it ensures high matching coverage~\cite{akbarpour2020thickness,baumler2022superiority} and provides a stable mechanism for collecting informative departure data before any data-driven policy switching is performed.
This conservative initialization avoids unstable or myopic decisions that could arise from insufficient early-stage observations.
In subsequent time windows, the system uses departure data collected in the preceding period to inform the next policy decision.
We assume a system environment with a stable or cyclically stable user distribution; addressing cold-start effects under abrupt structural changes is left for future work (See Section~\ref{sec:conc}).
Accordingly, the numerical results reported in Section~\ref{sec:num} exclude the initial learning phase and measure performance only after the system has reached a steady operating regime.

\section{Model and Methodology}
This section defines the formal mathematical environment and the matching policies that serve as the foundation for our $\mathsf{Hybrid}$ framework. We subsequently detail the performance metrics and the statistical learning architecture required to implement the data-driven policy decision.

\subsection{Matching Model Setting and Policies}

We consider a continuous-time dynamic matching market following the standard framework in \cite{akbarpour2020thickness,anderson2017efficient,baumler2022superiority}. We analyze a continuous-time matching market operating over a finite time horizon $[0,T]$, where $T>0$ denotes the end of the observation period. 
Agents (users) arrive according to a Poisson process with rate $\lambda$ and join the pool $Z_t$ of currently unmatched agents.  
Compatibility between any two agents is probabilistic, represented by independent Bernoulli trials with success probability $p$, yielding an effective density $d = p\cdot\lambda$.  
Each agent~$v_i$ is endowed with a maximum sojourn time $X_i$, independently sampled from a departure distribution.  
An agent exits the market either through a successful match with a compatible partner or by perishing when their sojourn time elapses, an event we refer to as departure.
This setting naturally induces a core trade-off between matching agents immediately and delaying matches to thicken the market.
Two canonical static policies have been extensively studied as performance benchmarks:

\paragraph{$\mathsf{Greedy}$ policy.}
Upon arrival, an agent is immediately matched to a compatible partner currently in the pool, chosen uniformly at random.  
If no such partner exists, the agent remains in the pool to wait for future arrivals.  
This policy prioritizes immediacy and commonly achieves high short-run match rates.

\paragraph{$\mathsf{Patient}$ policy.}  
Under the $\mathsf{Patient}$ policy, matching is deferred until an agent becomes critical at time $t+X_i$.  
A critical agent attempts to match with a compatible partner from the pool; if none exists, the agent perishes.  
By waiting, this policy aims to thicken the market and mitigate premature mismatches.

\vspace{3mm}

Although the $\mathsf{Greedy}$ and $\mathsf{Patient}$ policies perform well in different market regimes, with $\mathsf{Patient}$ achieving higher market thickness and matching efficiency, and $\mathsf{Greedy}$ incurring lower waiting costs and congestion, neither policy dominates across all performance metrics.
\cite{baumler2022superiority} shows that the trade-off between platform matching efficiency and agents’ waiting costs critically depends on the distribution of user departures.
In practice, real-world matching markets, such as hospitals, call centers, resource allocation platforms, or online marketplaces, exhibit time-varying departure patterns and dynamic behavioral uncertainty.
Consequently, market density and departure risk can fluctuate substantially across time windows, rendering fixed matching policies inherently inflexible.
This motivates the development of a data-driven mechanism capable of adapting the matching strategy in real time.

\paragraph{$\mathsf{Hybrid}$ framework.}
The $\mathsf{Hybrid}$ framework adaptively selects between the $\mathsf{Greedy}$ and $\mathsf{Patient}$ policies based on real-time data.  
The system operates over decision windows of size $w$.  
In each window, the platform aggregates behavioral and operational information from the previous window to estimate the departure-time distribution, which then informs the selection of the preferred matching policy for the subsequent window.

\vspace{3mm}

Through this repeated loop of estimation, policy selection, execution, and feedback adjustment, the $\mathsf{Hybrid}$ framework forms a closed-loop learning mechanism that continuously adapts to evolving market conditions.  
This enables the system to dynamically balance the benefits of platform matching efficiency and agents’ waiting costs.

\subsection{System Performance Metrics}

One of the key objectives of the matching system is to maximize the number of successful matches, which is equivalent to minimizing the number of agents who fail to match and ultimately depart the system. 
We refer to this quantity as \textit{Loss}, and it serves as a central measure of the system’s matching efficiency. For matching policy $\pi \in \{\mathsf{Greedy}, \mathsf{Patient}, \mathsf{Hybrid}\}$, The loss $\mathcal{L}$ is defined as:
\begin{equation*}
\mathcal{L}_{\pi}=\frac{\mathbb{E}[A_{[0,T]}-M_{[0,T]}-Z_{T}]}{A_{[0,T]}}.
\label{eq:loss}
\end{equation*}
Here, $A_{[0,T]}$ and $M_{[0,T]}$ denote the cumulative numbers of agent arrivals and successful matches over $[0,T]$, respectively, while $Z_T$ represents the number of agents remaining unmatched in the pool at time $T$.
Beyond matching efficiency, system designers also care about user-experience–related dimensions, such as expected waiting time and the level of market congestion. 
We define \textit{Congestion} as the average size of the unmatched pool during the matching horizon; lower congestion indicates a less crowded market environment, implying shorter search delays and a smoother operational workload for the system.
Our $\mathsf{Hybrid}$ framework is designed to improve overall system performance by jointly considering matching efficiency, user waiting experience, and market congestion. 
It aims to outperform the two canonical static policies, $\mathsf{Greedy}$ and $\mathsf{Patient}$, by achieving a more balanced trade-off across these metrics.
Within the $\mathsf{Hybrid}$ framework, policy switching between $\mathsf{Patient}$ and $\mathsf{Greedy}$ at each decision window is governed solely by the estimated loss threshold. 
An ideal outcome of the $\mathsf{Hybrid}$ framework is that it sacrifices only a marginal amount of matching efficiency, i.e., loss increases only slightly, in exchange for substantial reductions in both waiting time and congestion. 
Such a trade-off is meaningful in practice, as even a small decrease in system-level match rate can yield disproportionately large gains in user experience and operational stability. 
This makes the hybrid policy particularly attractive from an applied perspective, offering a more balanced and practically relevant performance profile than either static benchmark strategy.

\subsection{Statistical Learning Model}

We utilize a computationally efficient Statistical Learning Model (SLM) to estimate the agent departure time distribution. 
We specifically model the departure time using a Log-Normal distribution, defined by parameters $(\mu, \sigma)$.
This choice is motivated by the distribution's inherent non-negativity and its ability to capture the positive skewness (long-tail) characteristic commonly observed in waiting time data, where most agents depart quickly but a few remain patient for extended periods.
The estimated parameters $\mathbf{x} = [\mu, \sigma]^\top$ serve as the input to a Multi-Layer Perceptron (MLP). 
This MLP acts as a non-linear function approximator, establishing the mapping between the estimated distribution parameters $\mathbf{x}$ and the decision.

The core of our learning-based framework is a predict-then-decide pipeline that maps observed agent behavior to an optimal matching policy. This process follows a specific sequence. 
\begin{enumerate}
    \item \textit{Calibration and Training}: System administrators (e.g., hospital managers) first define a loss tolerance threshold $\tau$. Using this $\tau$ and historical or simulated data, we train a binary classifier, implemented via an MLP, to learn the optimal policy boundary across the parameter space $(\mu, \sigma)$ of the log-normal departure distribution.
    \item \textit{Estimation}: During real-time operation, at the end of each decision window $w$, the system collects departure data from the preceding period (see Section~\ref{subsec:alg} for details). We use point estimation to determine the current distribution parameters $(\mu_t, \sigma_t)$.
    \item \textit{Adaptive Decision}: These estimated parameters are fed into the pre-trained classifier to determine the matching policy for the subsequent window.
\end{enumerate}

\subsection{Policy Decision Model}

We formulate the policy selection problem as a binary classification task. 
The objective is to determine whether a $\mathsf{Patient}$ policy or a $\mathsf{Greedy}$ policy is preferred for the current system state, depends on the estimated departure parameters $\mathbf{x}= [\mu, \sigma]^\top$.
\begin{definition}
The loss tolerance threshold $\tau$
 quantifies the permissible performance gap between the proposed hybrid framework and the static $\mathsf{Patient}$ policy.
 \label{def}
\end{definition}
Theorem~\ref{thm:lb} and \ref{thm:wt} provide the asymptotics of the loss and waiting time for static matching policies, theoretically guaranteeing the comparability of the static $\mathsf{Patient}$ policy, the $\mathsf{Greedy}$ policy, and our $\mathsf{Hybrid}$ framework.
\begin{theorem}[Theoretical limits and policy behavior]
The following proposition rephrases Theorems 4.10 and 4.12 in \cite{baumler2022superiority} using our notation.
For any departure-time distribution with finite mean, the loss of any admissible policy is bounded below by an exponential rate, i.e., $\Omega(e^{-d})$.
The $\mathsf{Patient}$ policy achieves loss $\Theta(e^{-d})$ under both unit and exponential departure distributions.
The $\mathsf{Greedy}$ policy exhibits distributional sensitivity: 
While it achieves the $\Theta(e^{-d})$ loss under unit departure times, its performance deteriorates to only $\Theta(1/d)$ under exponential departure times.

\label{thm:lb}
\end{theorem}
Theorem~\ref{thm:lb} highlights the inherent sensitivity of the $\mathsf{Greedy}$ policy to the underlying departure-time distribution.
By adopting a log-normal family, we model a continuous spectrum of departure uncertainty that interpolates between the unit and exponential cases, thereby illustrating how performance degrades as variability increases.
Having established this distributional fragility in terms of loss, we next turn to agents’ waiting times, as characterized in Theorem~\ref{thm:wt}.

\begin{theorem}[Asymptotic waiting time of $\mathsf{Greedy}$ policy]
This theorem restates Theorem~4.14 in \cite{baumler2022superiority} in our notation.
For any departure-time distribution, the average waiting time under the $\mathsf{Greedy}$ policy is upper bounded by $O(1/d)$.
 
\label{thm:wt}
\end{theorem}
\noindent
Neither policy uniformly dominates the other: while $\mathsf{Patient}$ offers robustness in matching efficiency, $\mathsf{Greedy}$ is better suited for minimizing waiting times.

\begin{figure}[t]
\centering
\includegraphics[width=0.6\textwidth]{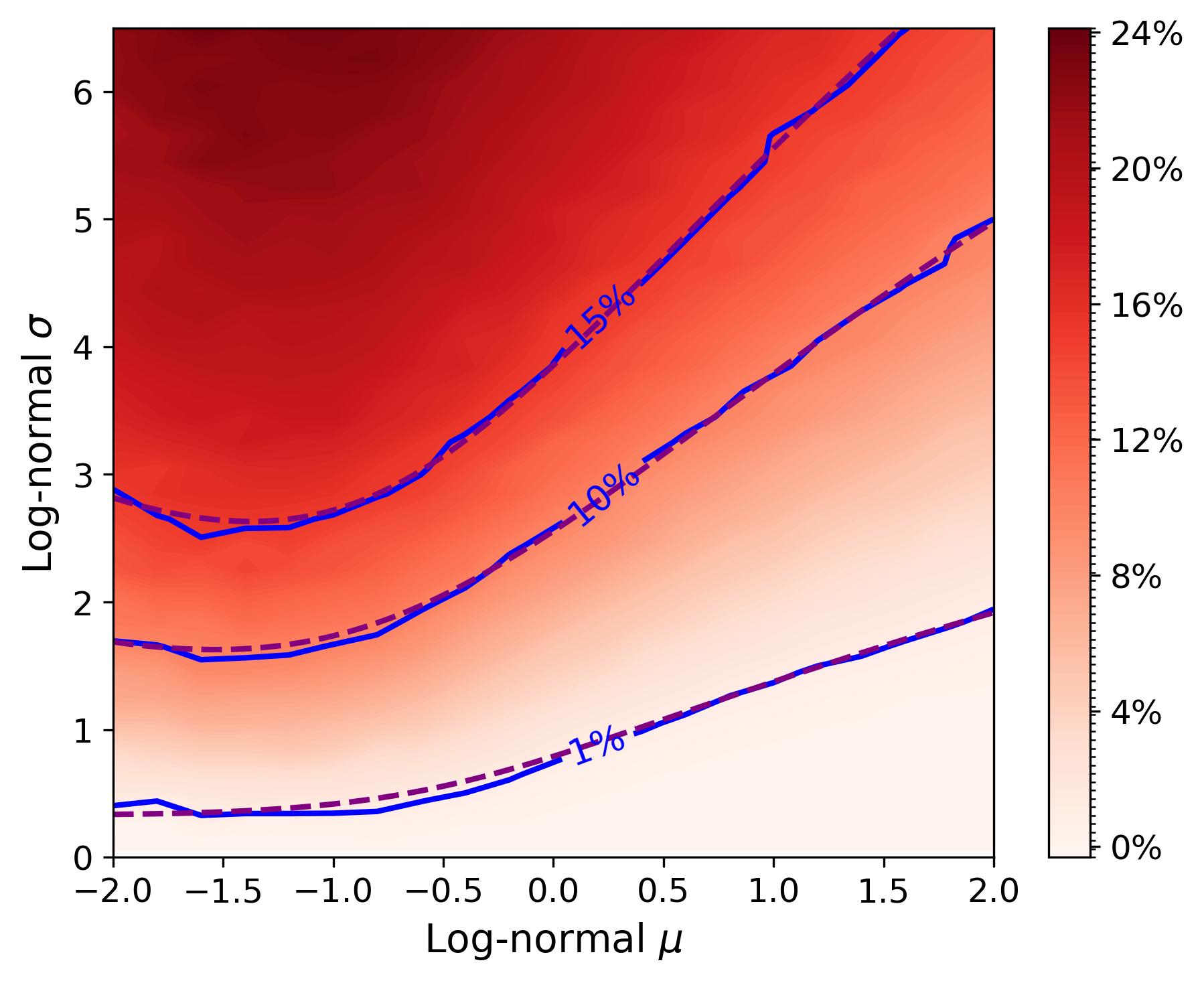}
\caption{Heatmap of $\mathsf{Greedy}$ vs. $\mathsf{Patient}$ policy gap across Log-normal parameters ($\mu$, $\sigma$). Solid lines trace actual contours for loss tolerance threshold $\tau\in\{1\%, 10\%, 15\%\}$.; dashed lines show framework-fitted decision thresholds.}\label{heatmap}
\end{figure}

Figure~\ref{heatmap} illustrates the loss difference between two pure policies, $\frac{\mathcal{L}_{\mathsf{Greedy}}}{\mathcal{L}_\mathsf{Patient}} - 1$, across the Log-Normal parameter space $(\mu, \sigma)$. 
The shaded areas represent this loss difference, clearly revealing a non-linear decision boundary. 
The darker red area indicates regions where this ratio is significantly greater than zero (i.e., $
\mathcal{L}_{\mathsf{Greedy}} > \mathcal{L}_{\mathsf{Patient}}$), signifying that the $\mathsf{Patient}$ policy yields higher matching efficiency due to its substantially lower loss compared to the $\mathsf{Greedy}$ policy.
The operational decision boundary is established by setting the loss difference to a fixed critical value (e.g., 10\%), which reflects practical trade-offs, such as the cost of monitoring agents. 
The close alignment between the fitted (dashed) and critical (solid) lines demonstrates the effectiveness of the MLP as a non-linear function approximator, accurately capturing the complex optimal policy boundary.

Rather than estimating the distribution itself, the model functions as a policy selector. It estimates the relative performance gap between the $\mathsf{Greedy}$ and $\mathsf{Patient}$ policies. Specifically, we define a performance score $\varsigma(\mu_t, \sigma_t)$ that represents the normalized loss difference:
\begin{equation*}
\varsigma(\mu_{t}, \sigma_{t})=\frac{\mathcal{L}_{\mathsf{Greedy}}(\mu_{t}, \sigma_{t})}{\mathcal{L}_{\mathsf{Patient}}(\mu_{t}, \sigma_{t})} -1.
\end{equation*}
The classifier is trained to predict whether this score $\varsigma$ exceeds a loss tolerance threshold $\tau$. The final decision rule applied at each operational stage is defined as:
\[
\text{Policy} \leftarrow
\begin{cases}
\mathsf{Patient} & \text{if } \varsigma(\mu_{t}, \sigma_{t})\geq \tau, \\
\mathsf{Greedy} & \text{otherwise}
\end{cases}
\] 
The decision threshold $\tau$ can be flexibly adjusted to accommodate asymmetric operational costs. 
This data-driven approach enables computational efficiency during deployment.

\subsection{Heuristic Algorithm for $\mathsf{Hybrid}$}
\label{subsec:alg}
We propose a $\mathsf{Hybrid}$ decision algorithm (Algorithm~\ref{alg:two}) that dynamically adapts the matching strategy based on the current system state.
\RestyleAlgo{ruled}
\begin{algorithm}[hbt!]
\caption{$\mathsf{Hybrid}$ Decision Algorithm}\label{alg:two}
\KwIn{Updated decision threshold $\tau$. 
Updated decision window size $w$. Estimation and classifier model.
}
\KwOut{Updated policy.}

Initialize empty agent batch $\mathcal{A}$.

\For{$t \leftarrow 0$ \KwTo $T$}
{

\If{\text{Agent Arrives}}{

Acquire agent information $A_i$,

Store $A_i$ in batch $\mathcal{A}$.
}
\eIf{$t$ $\text{mod}$ $w$ $==0$}{

    Estimate the Log-normal parameters $(\mu_{t}, \sigma_{t})$ from collected departure times of agents during the period.

    Calculate $\varsigma(\mu_{t}, \sigma_{t})$.  

    \eIf{$\varsigma(\mu_{t}, \sigma_{t}) \geq \tau$}{
        Policy $\leftarrow$ $\mathsf{Patient}$,
    }{
        Policy $\leftarrow$ $\mathsf{Greedy}$.
    }
    Reset agent batch $\mathcal{A}$.
    }
    {
    Policy remains unchanged.
    }
}
\end{algorithm}
This heuristic approach leverages the MLP's computationally efficient prediction score and integrates it with two critical control parameters: 
the decision threshold $\tau$ and the window size $w$. 
The algorithm indicates that policy determination is not static but periodically updated at every $w$ time steps during the operational flow.

\section{Numerical Study}
\label{sec:num}
We employ a continuous-time, event-driven simulation framework, which initializes with an empty pool at time $T=0$. 
Agents arrive according to a Poisson process with rate $\lambda$. 
Upon arrival, each agent is assigned an independent departure time, drawn from a specified distribution. 
The active matching policy is then applied to identify compatible pairs and remove selected matched agents from the pool.
To evaluate steady-state system performance and mitigate initialization bias, we simulate the market over a total horizon of $T$ and record performance metrics only after a mixing period of $T_0$.
The arrival rate is set to a representative value of $\lambda$, consistent with the scale of large real-world matching markets studied in the literature \cite{baccara2014child,unver2010dynamic}.
Our proposed $\mathsf{Hybrid}$ framework depends on two deterministic control parameters: the decision threshold $\tau$ and the estimation window size $w$ (batch). 
In practice, administrators (e.g., hospital managers) would calibrate these parameters to align with specific operational objectives.

This Section evaluates the proposed $\mathsf{Hybrid}$ framework by comparing its performance against static benchmarks: the $\mathsf{Greedy}$ policy and the $\mathsf{Patient}$ policy. 
We conduct a comprehensive ablation study to dissect the framework's performance and understand its inner workings.
In Section~\ref{subsec:threshold}, we analyze the sensitivity of system performance to variations in the loss tolerance threshold $\tau$.
In Section~\ref{subsec:ws}, we investigate the impact of the estimation window size $w$ on overall system performance.
In Section~\ref{subsec:ratio}, we explore the internal composition of the $\mathsf{Hybrid}$ framework by examining the ratio between the two constituent policies ($\mathsf{Greedy}$ vs. $\mathsf{Patient}$) under different conditions.

\subsection{Loss Tolerance Threshold}
\label{subsec:threshold}
\begin{figure}[tbp]
    \centering
    \begin{subfigure}[b]{0.325\textwidth}
        \centering
        \includegraphics[width=\textwidth]{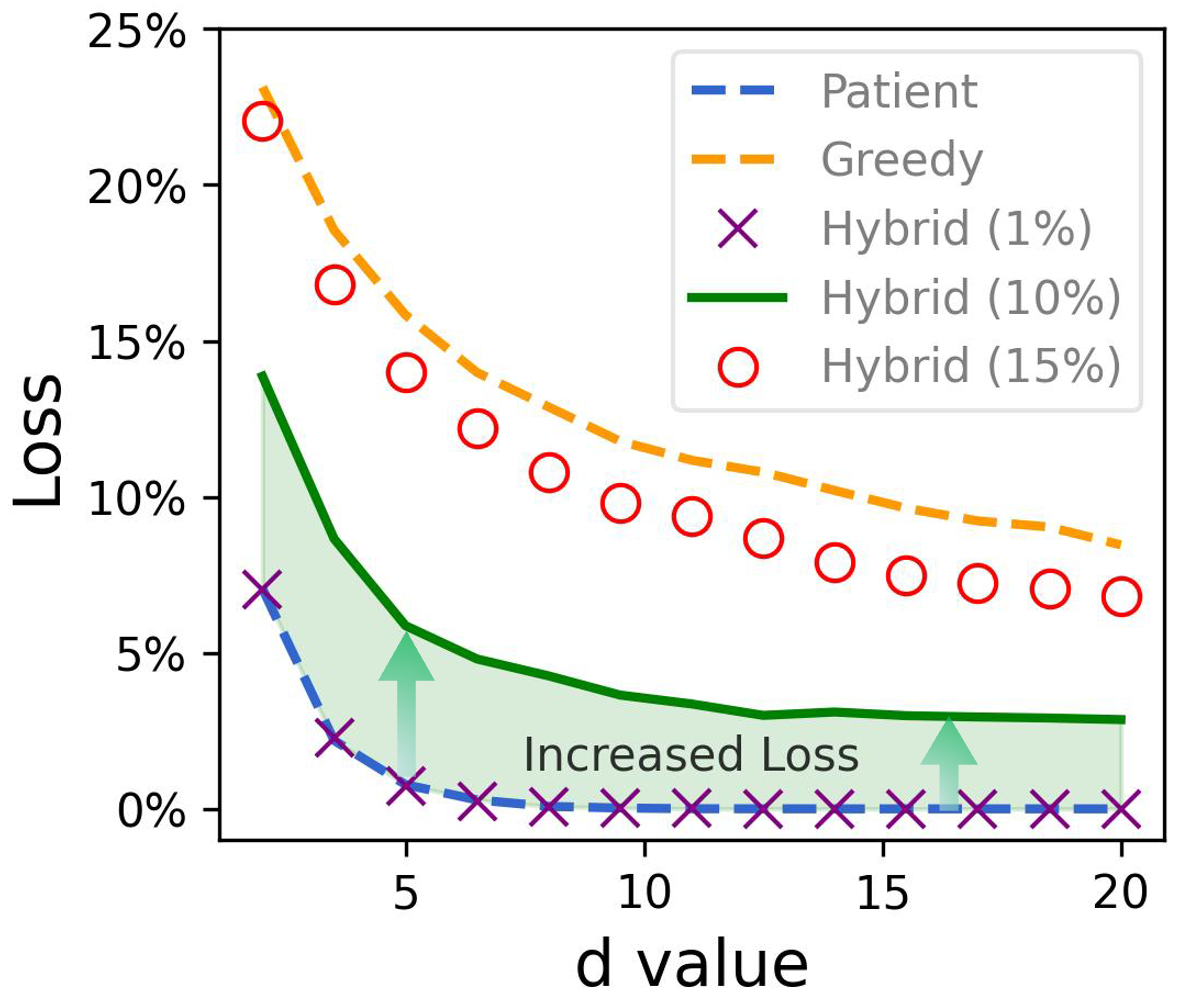}
        \caption{Loss}
        \label{loss_threshold_d}
    \end{subfigure}
    \begin{subfigure}[b]{0.325\textwidth}
        \includegraphics[width=\textwidth]{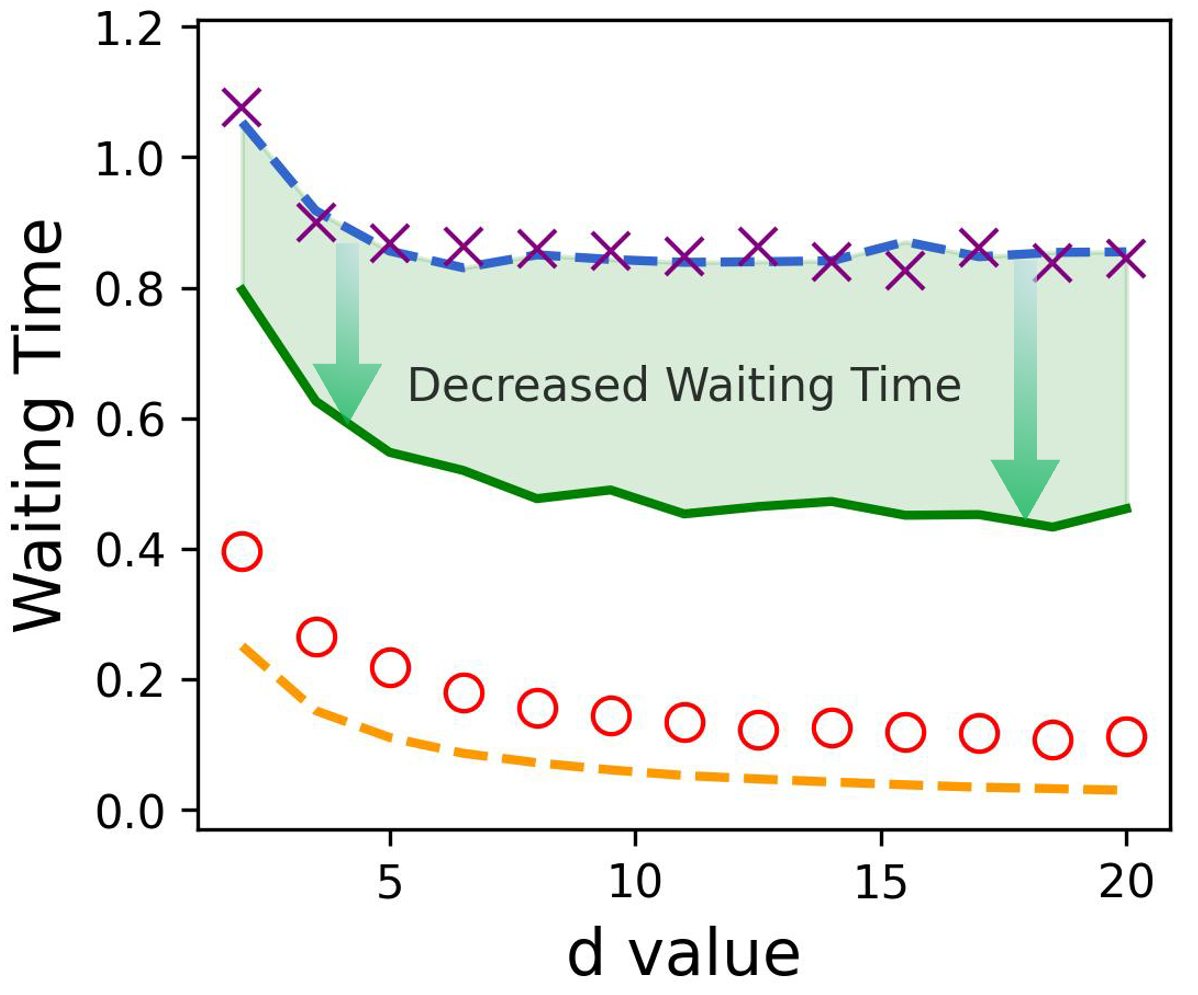}
        \caption{Expected Waiting Time}
        \label{waiting_time_threshold_d}
    \end{subfigure}
    \begin{subfigure}[b]{0.325\textwidth}
        \centering
        \includegraphics[width=\textwidth]{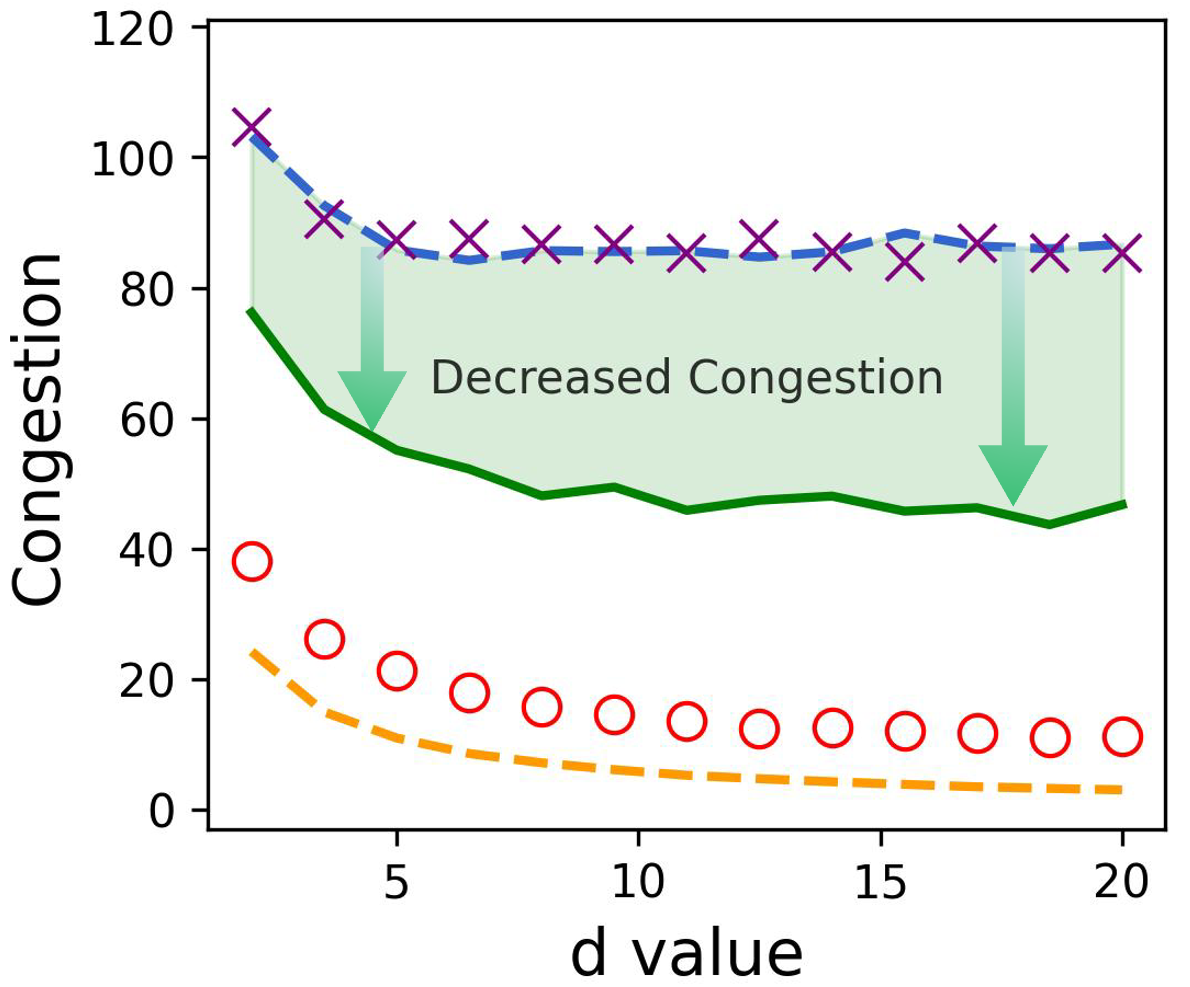}
        \caption{Congestion}
        \label{pool_size_threshold_d}
    \end{subfigure}
    
    \caption{Performance of the $\mathsf{Hybrid}$ framework (under varying thresholds $\tau$) versus static policies as $d$ varies.
    Fixed parameters include $w=0.3$, $\lambda=100$, $T_0=50$, and $T=100$. Results are averaged over $k=10$ independent runs.}
    \label{performance_threshold_d}
\end{figure}

The Loss tolerance threshold ($\tau$) serves as the critical decision boundary within our $\mathsf{Hybrid}$ framework, dictating the switch between the $\mathsf{Patient}$ and $\mathsf{Greedy}$ policies. 
Intuitively, it represents the system's tolerance level for loss:
If the predicted loss is acceptable, the system prioritizes the $\mathsf{Greedy}$ policy, emphasizing matching speed. 
Conversely, if the predicted loss is unacceptable, it executes the $\mathsf{Patient}$ policy, emphasizing matching coverage.
Therefore, setting this threshold directly reflects the system's strategic priority: 
A higher threshold signifies higher intolerance to loss, prioritizing high match coverage even at the expense of increased waiting times. 
Conversely, a lower threshold indicates a greater emphasis on cost and speed, accepting a slightly higher loss to ensure swift matching.
System density ($d$) is a key metric measuring the activity level or the saturation of potential matching opportunities within the market. 
As shown in Figure~\ref{performance_threshold_d}, we set different values of $d$ for observing the performance of policies. 

\begin{result}
The $\mathsf{Hybrid}$ framework demonstrates its flexibility and tunability by effectively interpolating between the $\mathsf{Patient}$ and $\mathsf{Greedy}$ static policies, enabling precise performance control across the entire operational spectrum defined by these two policies.
\label{result:threshold}
\end{result}

Result~\ref{result:threshold} highlights the tunability and interpolation capability of the $\mathsf{Hybrid}$ framework. 
As shown in Figure~\ref{performance_threshold_d}, system performance is bounded by two static extremes: 
The $\mathsf{Patient}$ policy, which prioritizes minimizing matching loss, and the $\mathsf{Greedy}$ policy, which prioritizes reducing waiting time and congestion. 
By varying the threshold $\tau$, the $\mathsf{Hybrid}$ policy operates continuously within the performance region spanned by these two benchmarks.
Figure~\ref{loss_threshold_d} shows that matching loss decreases either exponentially or inverse-linearly with system density $d$, reflecting the higher likelihood of compatible pairs in denser markets. 
This behavior is consistent with the theoretical prediction in Theorem~\ref{thm:lb}. 
Increasing the threshold (e.g., from 1\% to 10\% or 15\%) shifts the $\mathsf{Hybrid}$ policy toward the $\mathsf{Greedy}$ regime, thereby placing greater emphasis on cost efficiency.
Figures~\ref{waiting_time_threshold_d} and~\ref{pool_size_threshold_d} report the corresponding effects on waiting time and system congestion (pool size). 
These two metrics exhibit nearly identical trends, consistent with the fact that total waiting time is determined by the time integral of the pool size. As established in Theorem~\ref{thm:wt}, the waiting time for the  $\mathsf{Greedy}$ policy exhibits an inverse-linear trend. However, numerical observations reveal that the $\mathsf{Patient}$ policy results in constant waiting times.

The green shaded regions in Figure~\ref{performance_threshold_d} visualize the strategic trade-offs managed by the framework. 
Moving away from the $\mathsf{Patient}$ policy incurs additional matching loss (Figure~\ref{loss_threshold_d}), but yields substantial reductions in waiting time and congestion (Figures~\ref{waiting_time_threshold_d} and~\ref{pool_size_threshold_d}). 
This interpolation is highly non-linear: 
An intermediate threshold, such as $\tau=10\%$, captures most of the waiting-time and congestion reduction, approaching the performance of the $\mathsf{Greedy}$ policy, while incurring only a modest increase in loss relative to the $\mathsf{Patient}$ policy.
The underlying intuition is that allowing a small, controlled increase in matching loss prevents the accumulation of congestion and excessive waiting. 
As system density $d$ increases, higher natural compatibility enables the $\mathsf{Hybrid}$ framework to adopt larger thresholds $\tau$, accelerating matching speed while preserving protection against the extreme losses associated with a static $\mathsf{Greedy}$ policy.

\subsection{Window Size}
\label{subsec:ws}
The window size ($w$) represents the length of a decision batch in time.
Within each window, the system commits to a fixed matching policy and updates its policy only at window boundaries, based on the user departure distribution learned from data observed in the previous window.
During the initial window, the system defaults to the $\mathsf{Patient}$ policy.
At the end of each window, data collected during that period are used to estimate the distribution parameters $(\mu,\sigma)$ and determine the policy to be implemented in the subsequent window.
\begin{figure}[tbp]
    \centering
    \begin{subfigure}[b]{0.325\textwidth}
        \centering
    \includegraphics[width=\textwidth]{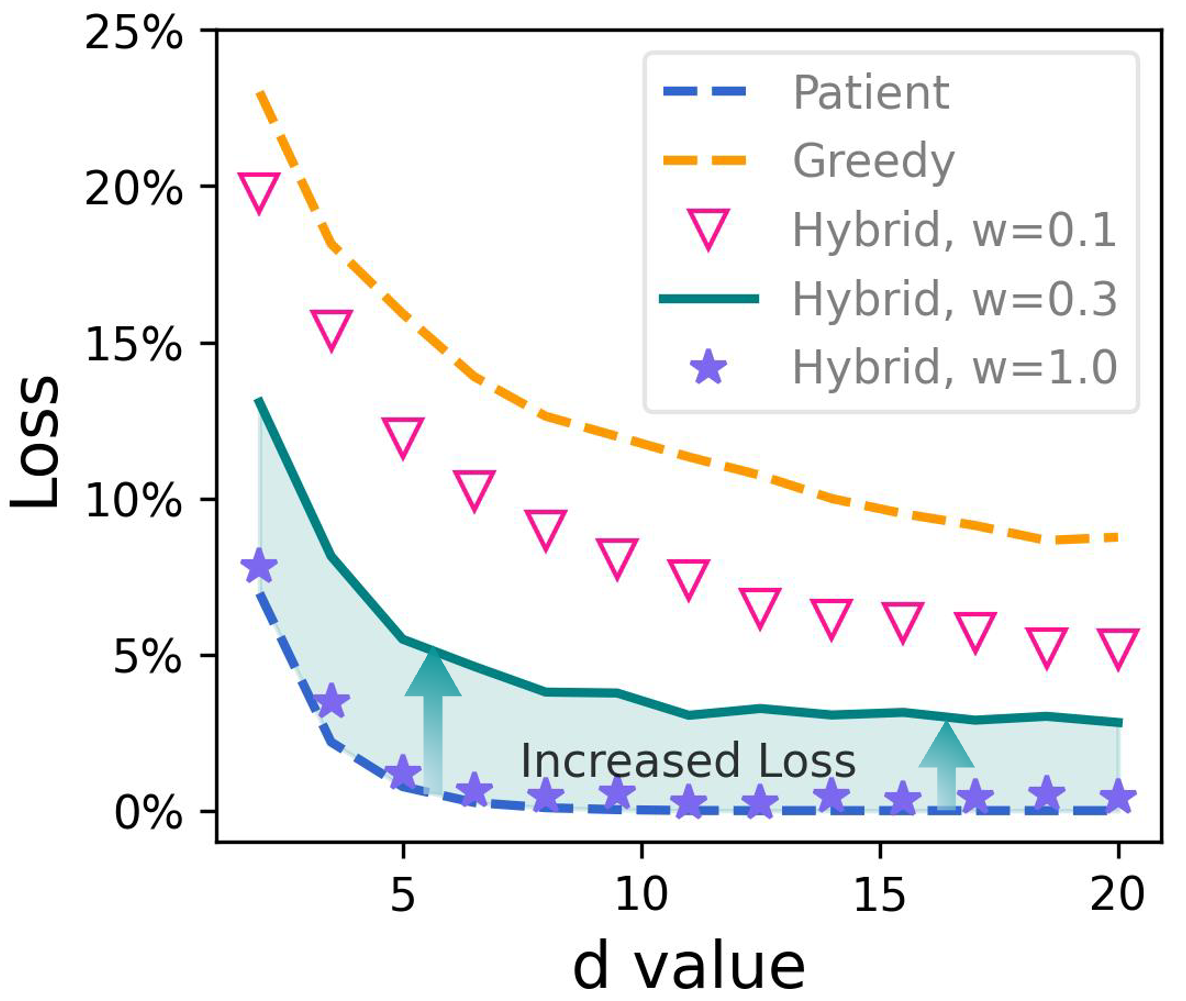}
        \caption{Loss}
        \label{loss_window_size_d}
    \end{subfigure}
    \begin{subfigure}[b]{0.325\textwidth}
        \includegraphics[width=\textwidth]{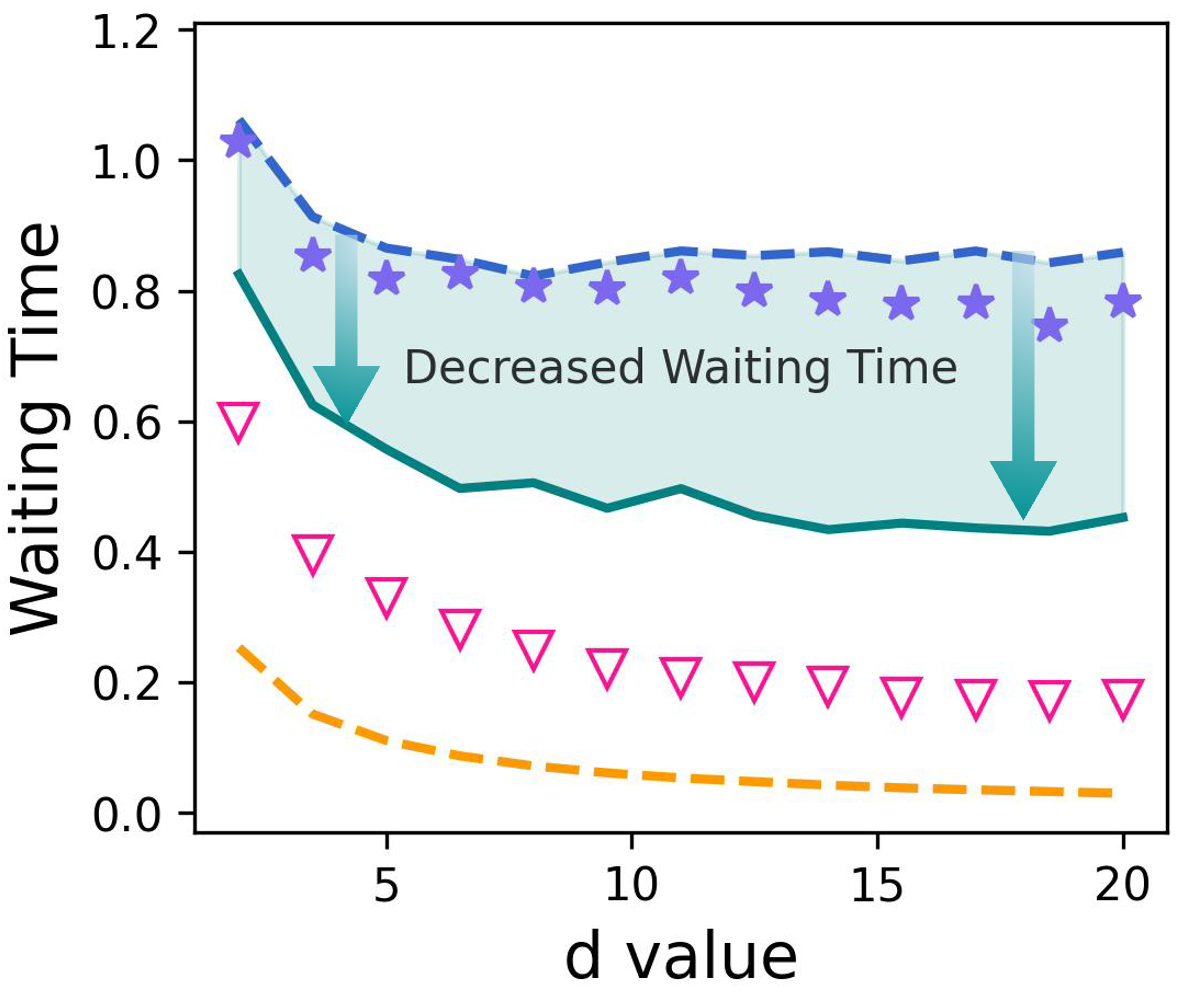}
        \caption{Expected Waiting time}
     \label{waiting_time_window_size_d}
    \end{subfigure}
    \begin{subfigure}[b]{0.325\textwidth}
        \centering
    \includegraphics[width=\textwidth]{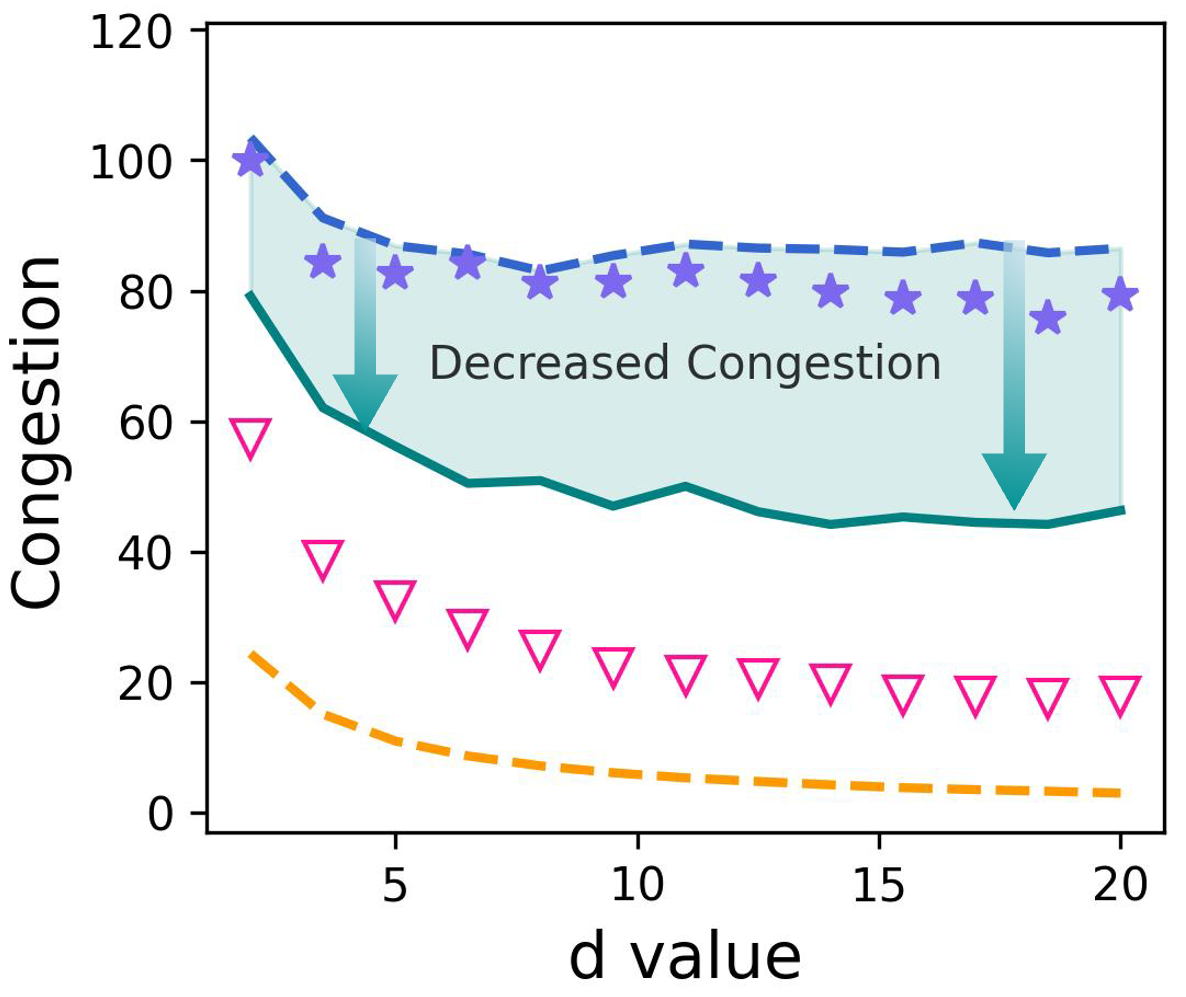}
        \caption{Congestion}
        \label{pool_size_window_size_d}
    \end{subfigure}
    \caption{Performance of the $\mathsf{Hybrid}$ framework (under varying window sizes $w$) versus static policies as $d$ varies.
    Fixed parameters include $\tau=10\%$, $\lambda=100$, $T_0=50$, $T=100$, and $k=10$.}
    \label{performance_windowsize_d}
\end{figure}
\begin{result}
Both the decision threshold $\tau$ and the window size $w$ serve as effective tuning parameters for the $\mathsf{Hybrid}$ framework.
These parameters enable consistent control over the same performance trade-off between loss, waiting time, and congestion.
Adjusting either parameter induces qualitatively consistent trade-off patterns, whereby increases in loss are accompanied by corresponding decreases in waiting time and congestion, relative to static benchmark policies.
\end{result}
This phenomenon is clearly illustrated in Figures~\ref{performance_threshold_d} and Figure~\ref{performance_windowsize_d}.
Figure~\ref{performance_threshold_d} shows that varying the decision threshold $\tau$ shifts the $\mathsf{Hybrid}$ framework smoothly between the $\mathsf{Patient}$ and $\mathsf{Greedy}$ regimes.
As $\tau$ increases, the framework trades off higher loss for lower waiting time and congestion, with consistent trends observed across different values of $d$.
Similarly, Figure~\ref{performance_windowsize_d} demonstrates that changing the window size $w$ produces qualitatively similar trade-off patterns:
increasing or decreasing $w$ leads to systematic shifts between loss and delay-related metrics, mirroring the effects induced by varying $\tau$.
The reason both parameters exhibit similar trade-off patterns is that they regulate the relative exposure of the system to the $\mathsf{Patient}$ and $\mathsf{Greedy}$ policies through different mechanisms.
While the loss tolerance threshold $\tau$ directly controls the system’s tolerance for loss in policy selection, the window size $w$ determines how frequently the system updates its policy based on newly learned user distributions.
Despite operating through distinct channels, both parameters effectively shift the policy mix in the same direction.

\subsection{Policy Switching Frequency}
\label{subsec:ratio}

\begin{figure}[tbp]
\centering
\includegraphics[width=0.8\textwidth]{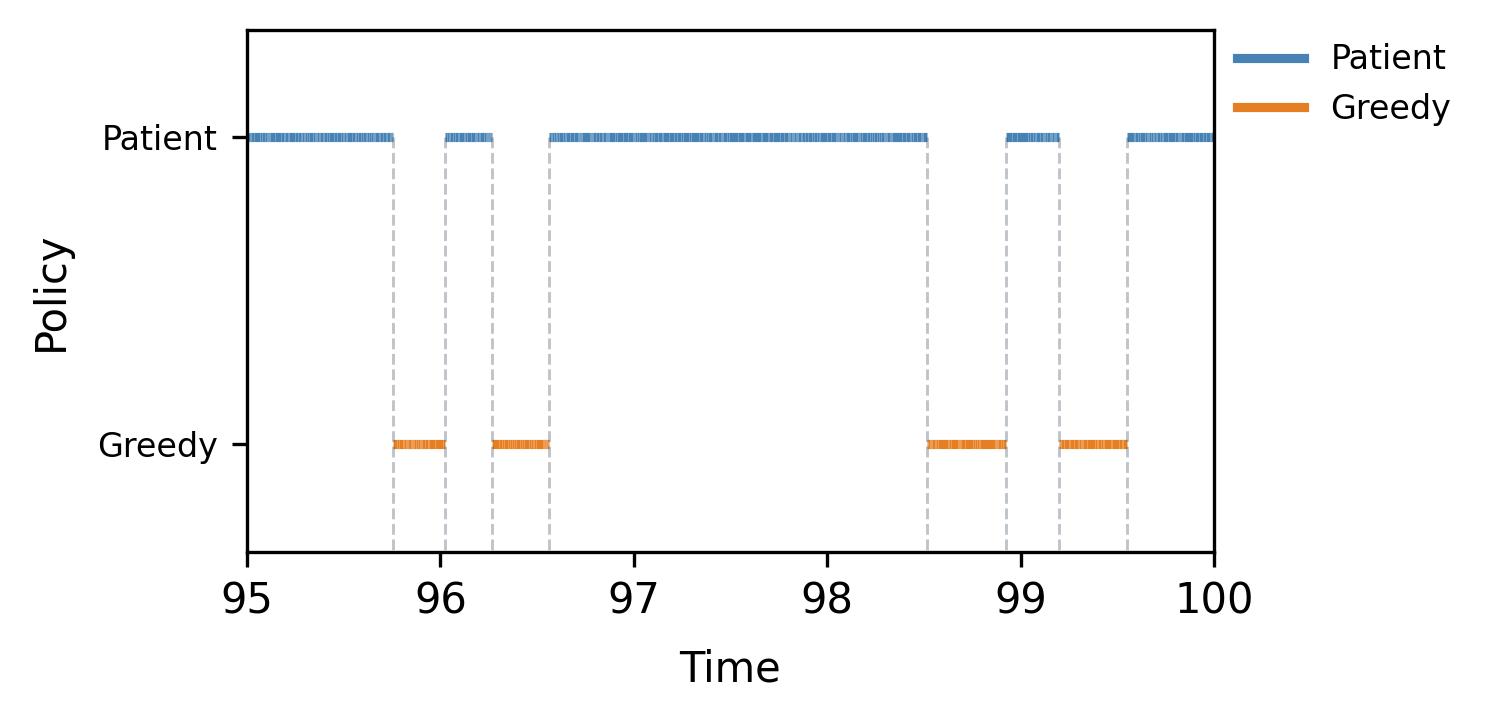}
\caption{Dynamic policy selection under the $\mathsf{Hybrid}$ framework with $\tau=10\%$, $w=0.3$, over the interval $T \in [95,100]$.}
\label{policy_list}
\end{figure}

We analyze the proportionate use of the $\mathsf{Greedy}$ versus $\mathsf{Patient}$ policies within the $\mathsf{Hybrid}$ framework.
Figure~\ref{policy_list} illustrates the sequence of policy choices made by the $\mathsf{Hybrid}$.
The figure shows that the framework does not commit to a single static policy, but instead alternates between the $\mathsf{Patient}$ and $\mathsf{Greedy}$ policies over time.
While the $\mathsf{Patient}$ policy is selected more frequently, the $\mathsf{Greedy}$ policy is periodically activated, resulting in discrete policy switches rather than continuous oscillations.
The intermittent activation of the $\mathsf{Greedy}$ policy suggests that the framework selectively deploys fast matching when system conditions warrant immediate congestion relief.
\begin{result}
The window size $w$ directly affects policy switching frequency: smaller windows lead to more frequent and regular alternation between the $\mathsf{Patient}$ and $\mathsf{Greedy}$ policies, while larger windows result in increasingly stable dominance by a single policy.

\end{result}
\begin{figure}[tbp]
\centering
\includegraphics[width=0.7\textwidth]{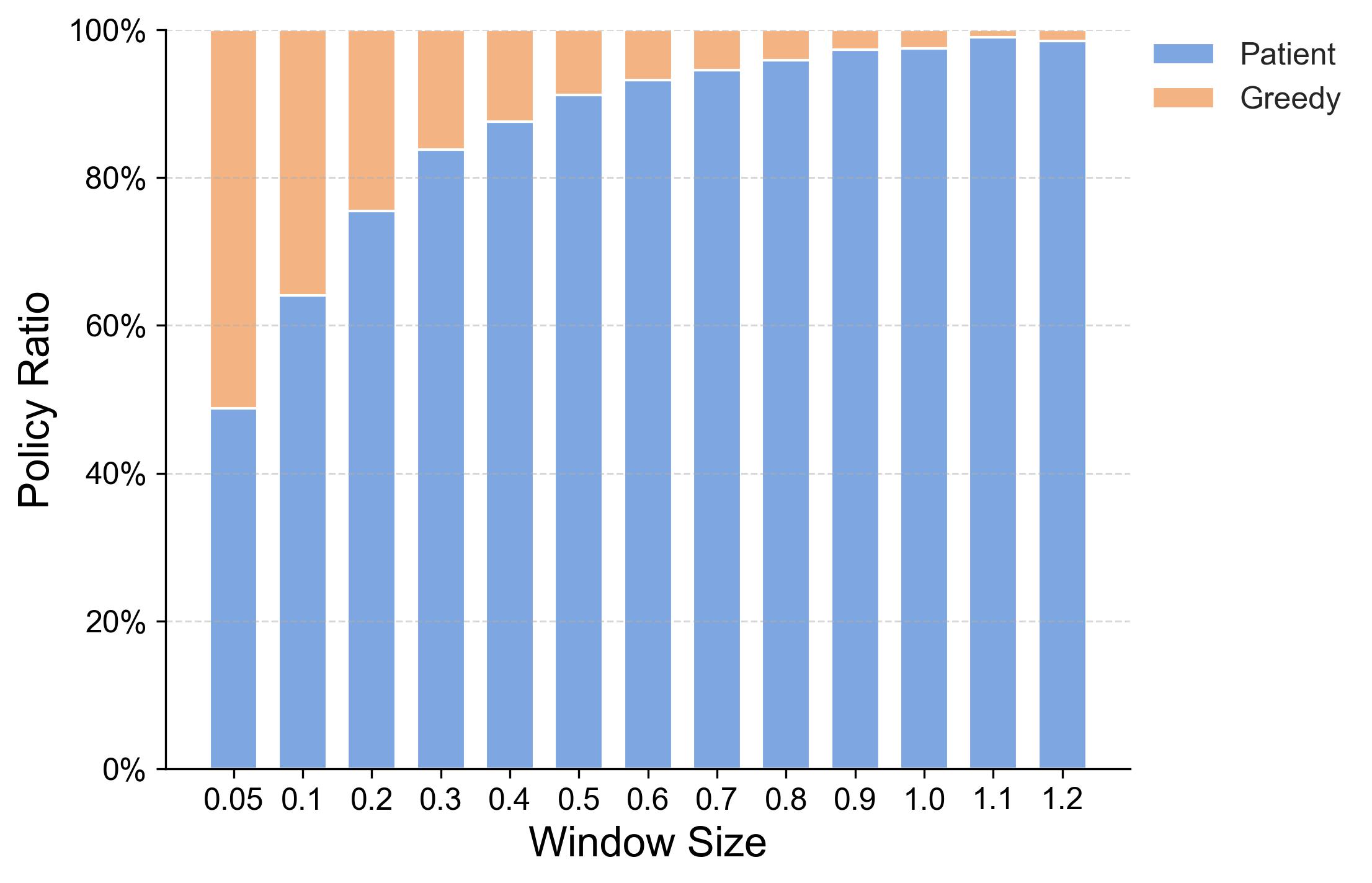}
\caption{Policy frequency under the $\mathsf{Hybrid}$ framework as window size varies, $\tau=10\%$ and $k=10$, over the interval $T \in [80,100]$.}
\label{policy_ratio}
\end{figure}

Figure~\ref{policy_ratio} shows how the policy usage ratio evolves as the window size varies.
For small window sizes, both $\mathsf{Patient}$ and $\mathsf{Greedy}$ policies are activated with non-negligible frequency, reflecting frequent policy switching.
As the window size increases, the share of time spent under the $\mathsf{Patient}$ policy steadily rises, while activations of the $\mathsf{Greedy}$ policy become increasingly rare.
For sufficiently large windows, the policy usage becomes almost entirely dominated by $\mathsf{Patient}$, indicating a transition toward quasi-static behavior.
This pattern arises because smaller windows rely on shorter historical samples, making policy decisions more sensitive to local fluctuations in the estimated departure distribution.
Larger windows aggregate information over longer horizons, smoothing out short-term variability and yielding more stable estimates, which in turn consistently favor the policy that performs best on average.

\section{Conclusion and Discussions}\label{sec:conc}

In this paper, we develop and evaluate a learning-based hybrid decision framework for optimizing dynamic matching markets with uncertain user departure times. 
By modeling departure behavior using a log-normal distribution and employing a neural network based decision model, the $\mathsf{Hybrid}$ framework dynamically switches between \textsf{Patient} and \textsf{Greedy} matching policies in response to evolving market conditions. 
Through a continuous feedback loop involving the user, operational platform, and decision support analyst module, the system adapts to stochastic arrivals and departures in real time. 
Extensive simulation results demonstrate that the proposed framework effectively navigates the trade-off between matching efficiency and system cost, offering a level of flexibility that static single-policy approaches cannot achieve. 
In particular, our results show that high matching efficiency need not require prohibitive congestion or waiting costs. 
Rather than claiming a strict Pareto improvement, we find that the hybrid framework can sacrifice a small amount of matching efficiency to achieve a substantial reduction in user waiting time and market congestion, leading to significantly improved overall system performance in realistic operating regimes.

Based on the framework proposed in this paper, several natural extensions and open research questions arise that merit further investigation.

\begin{itemize}
\item 
An important direction concerns the causal relationship between input data and decision outcomes. 
Our framework determines the matching policy for the next time window based on market statistics observed in the previous period. 
As a result, the robustness of the matching outcomes is inherently linked to the stability of the state parameters of arriving users. 
Understanding the quantitative causal relationship between the stability of these input signals and the robustness of the resulting matching decisions is a nontrivial problem. 
A more explicit causal characterization could shed light on how estimation noise, delayed feedback, or structural changes in arrivals propagate through the decision pipeline and affect system performance.
\item 
The matching structure itself can be meaningfully extended beyond the current setting. 
For example, the framework can be generalized to dynamic matching mechanisms on bipartite or tripartite graphs~\cite{kakimura2021dynamic}. 
Under the theoretical insights developed in this work, the qualitative relationship between matching efficiency, user waiting time, and market congestion is expected to persist. 
However, richer structural factors would be introduced, leading to quantitative changes in the trade-off. 
In particular, asymmetries in matching power across different sides of a bipartite market, or the presence of intermediary agents in tripartite settings, may significantly alter the system dynamics and warrant dedicated analysis.
\item 
The framework naturally lends itself to hybrid optimization models based on multiple system performance metrics. 
Rather than optimizing matching efficiency alone, future work could consider composite objective functions that jointly account for matching efficiency, user waiting time, and congestion costs. 
This would motivate the development of new optimization algorithms tailored to such multi-criteria objectives. 
In addition, incorporating user heterogeneity into the model, such as differences in patience, urgency, or valuation, would further enhance the framework’s realism and allow the design of policies that explicitly account for heterogeneous user experiences.
\end{itemize}

\begin{credits}
\subsubsection{\ackname} 
This work was supported in part by the JSPS Grant-in-Aid for Transformative Research Areas (A) 22H05106, and the National Natural Science Foundation of China (Grants 72571132 and 72394363/72394360).
\end{credits}

% \subsubsection{\discintname}
% It is now necessary to declare any competing interests or to specifically
% state that the authors have no competing interests. Please place the
% statement with a bold run-in heading in small font size beneath the
% (optional) acknowledgments\footnote{If EquinOCS, our proceedings submission
% system, is used, then the disclaimer can be provided directly in the system.},
% for example: The authors have no competing interests to declare that are
% relevant to the content of this article. Or: Author A has received research
% grants from Company W. Author B has received a speaker honorarium from
% Company X and owns stock in Company Y. Author C is a member of committee Z.
% \end{credits}
% %
% % ---- Bibliography ----
% %
% % BibTeX users should specify bibliography style 'splncs04'.
% % References will then be sorted and formatted in the correct style.
% %
\bibliographystyle{splncs04}
\bibliography{mybib}

@article{akbarpour2020thickness,
  title={Thickness and information in dynamic matching markets},
  author={Akbarpour, Mohammad and Li, Shengwu and Gharan, Shayan Oveis},
  journal={Journal of Political Economy},
  volume={128},
  number={3},
  pages={783--815},
  year={2020}
}

@conference{baumler2022superiority,
    author = {B{\"a}umler, Johannes and Bullinger, Martin and Kober, Stefan and Zhu, Donghao},
    booktitle = {Proceedings of the 24th ACM
Conference on Economics and Computation},
    title = {Superiority of instantaneous decisions in thin dynamic matching markets},
    year = {2023},
    pages = {390}
}

@article{anderson2017efficient,
  title={Efficient dynamic barter exchange},
  author={Anderson, Ross and Ashlagi, Itai and Gamarnik, David and Kanoria, Yash},
  journal={Operations Research},
  volume={65},
  number={6},
  pages={1446-1459},
  year={2017}
}

@article{guidolin2022ethical,
  title={Ethical decision making during a healthcare crisis: a resource allocation framework and tool},
  author={Guidolin, Keegan and Catton, Jennifer and Rubin, Barry and Bell, Jennifer and Marangos, Jessica and Munro-Heesters, Ann and Stuart-McEwan, Terri and Quereshy, Fayez},
  journal={Journal of Medical Ethics},
  volume={48},
  number={8},
  pages={504-509},
  year={2022}
}

@article{dickerson2017multi,
  title={Multi-organ exchange},
  author={Dickerson, John P and Sandholm, Tuomas},
  journal={Journal of Artificial Intelligence Research},
  volume={60},
  pages={639-679},
  year={2017}
}

@article{bertsimas2013fairness,
  title={Fairness, efficiency, and flexibility in organ allocation for kidney transplantation},
  author={Bertsimas, Dimitris and Farias, Vivek F and Trichakis, Nikolaos},
  journal={Operations Research},
  volume={61},
  number={1},
  pages={73-87},
  year={2013}
}

@article{papalexopoulos2024reshaping,
  title={Reshaping national organ allocation policy},
  author={Papalexopoulos, Theodore and Alcorn, James and Bertsimas, Dimitris and Goff, Rebecca and Stewart, Darren and Trichakis, Nikolaos},
  journal={Operations Research},
  volume={72},
  number={4},
  pages={1475-1486},
  year={2024}
}

@article{ashlagi2021kidney,
  title={Kidney exchange: An operations perspective},
  author={Ashlagi, Itai and Roth, Alvin E},
  journal={Management Science},
  volume={67},
  number={9},
  pages={5455-5478},
  year={2021}
}

@article{carvalho2021robust,
  title={Robust models for the kidney exchange problem},
  author={Carvalho, Margarida and Klimentova, Xenia and Glorie, Kristiaan and Viana, Ana and Constantino, Miguel},
  journal={INFORMS Journal on Computing},
  volume={33},
  number={3},
  pages={861-881},
  year={2021}
}

@article{morse2024centralized,
  title={Centralized Scheduling of Nursing Staff: A Rapid Review of the Literature},
  author={Morse, Lisa and Duncan, Hillary and Apen, Lynette V and Reese, Karin and Crawford, Cecelia L},
  journal={Nursing Administration Quarterly},
  volume={48},
  number={4},
  pages={347-358},
  year={2024}
}

@article{demeester2010hybrid,
  title={A hybrid tabu search algorithm for automatically assigning patients to beds},
  author={Demeester, Peter and Souffriau, Wouter and De Causmaecker, Patrick and Vanden Berghe, Greet},
  journal={Artificial Intelligence in Medicine},
  volume={48},
  number={1},
  pages={61-70},
  year={2010}
}

@article{kerimov2025optimality,
  title={On the optimality of greedy policies in dynamic matching},
  author={Kerimov, S{\"u}leyman and Ashlagi, Itai and Gurvich, Itai},
  journal={Operations Research},
  volume={73},
  number={1},
  pages={560-582},
  year={2025}
}

@article{kerimov2024dynamic,
  title={Dynamic matching: Characterizing and achieving constant regret},
  author={Kerimov, S{\"u}leyman and Ashlagi, Itai and Gurvich, Itai},
  journal={Management Science},
  volume={70},
  number={5},
  pages={2799-2822},
  year={2024}
}

@article{gupta2024greedy,
  title={Greedy algorithm for multiway matching with bounded regret},
  author={Gupta, Varun},
  journal={Operations Research},
  volume={72},
  number={3},
  pages={1139-1155},
  year={2024}
}

@article{elmachtoub2022smart,
  title={Smart “predict, then optimize”},
  author={Elmachtoub, Adam N and Grigas, Paul},
  journal={Management Science},
  volume={68},
  number={1},
  pages={9--26},
  year={2022}
}

@article{mandi2024decision,
  title={Decision-focused learning: Foundations, state of the art, benchmark and future opportunities},
  author={Mandi, Jayanta and Kotary, James and Berden, Senne and Mulamba, Maxime and Bucarey, Victor and Guns, Tias and Fioretto, Ferdinando},
  journal={Journal of Artificial Intelligence Research},
  volume={80},
  pages={1623--1701},
  year={2024}
}

@article{donti2017task,
  title={Task-based end-to-end model learning in stochastic optimization},
  author={Donti, Priya and Amos, Brandon and Kolter, J Zico},
  journal={Advances in Neural Information Processing Systems},
  volume={30},
  year={2017}
}

@article{sadana2025survey,
  title={A survey of contextual optimization methods for decision-making under uncertainty},
  author={Sadana, Utsav and Chenreddy, Abhilash and Delage, Erick and Forel, Alexandre and Frejinger, Emma and Vidal, Thibaut},
  journal={European Journal of Operational Research},
  volume={320},
  number={2},
  pages={271--289},
  year={2025}
}

@article{feng2023framework,
  title={The framework of parametric and nonparametric operational data analytics},
  author={Feng, Qi and Shanthikumar, J George},
  journal={Production and Operations Management},
  volume={32},
  number={9},
  pages={2685--2703},
  year={2023}
}

@article{warwick2005choosing,
  title={Choosing a robustness tuning parameter},
  author={Warwick, Jane and Jones, MC},
  journal={Journal of Statistical Computation and Simulation},
  volume={75},
  number={7},
  pages={581--588},
  year={2005},
  publisher={Taylor \& Francis}
}

@article{baccara2014child,
  title={Child-adoption matching: Preferences for gender and race},
  author={Baccara, Mariagiovanna and Collard-Wexler, Allan and Felli, Leonardo and Yariv, Leeat},
  journal={American Economic Journal: Applied Economics},
  volume={6},
  number={3},
  pages={133--158},
  year={2014},
  publisher={American Economic Association 2014 Broadway, Suite 305, Nashville, TN 37203-2425}
}

@article{unver2010dynamic,
  title={Dynamic kidney exchange},
  author={{\"U}nver, M Utku},
  journal={The Review of Economic Studies},
  volume={77},
  number={1},
  pages={372--414},
  year={2010},
  publisher={Wiley-Blackwell}
}

@conference{kakimura2021dynamic,
    author = {Kakimura, Naonori and Zhu, Donghao},
    booktitle = {Proceedings of the 17th Conference on Web and Internet Economics},
    title = {Dynamic bipartite matching market with arrivals and departures},
    year = {2021},
    pages = {544}
}

@article{kao2009directed,
  title={Directed regression},
  author={Kao, Yi-hao and Roy, Benjamin and Yan, Xiang},
  journal={Advances in Neural Information Processing Systems},
  volume={22},
  year={2009}
}

% \begin{thebibliography}{2}
% \bibitem{Akbarpour2020}
% Akbarpour, M., Li, S., Gharan, S. O.: Thickness and information in dynamic matching markets. Journal of Political Economy \textbf{128}(3), 783--815 (2020)

% \bibitem{B2023}
% Bäumler, J., Bullinger, M., Kober, S., Zhu, D.: Superiority of instantaneous decisions in thin dynamic matching markets. 24th ACM Conference on Economics and Computation (2023). \doi{
% 10.48550/arXiv.2206.10287}

% \bibitem{ref_book1}
% Author, F., Author, S., Author, T.: Book title. 2nd edn. Publisher,
% Location (1999)

% \bibitem{ref_proc1}
% Author, A.-B.: Contribution title. In: 9th International Proceedings
% on Proceedings, pp. 1--2. Publisher, Location (2010)

% \bibitem{ref_url1}
% LNCS Homepage, \url{http://www.springer.com/lncs}, last accessed 2023/10/25
% \end{thebibliography}
\end{document}